\documentclass[sigconf]{acmart}

\usepackage{tikz}
\usepackage{array}
\usepackage{tabularx}
\usepackage[acronym]{glossaries}
\makeglossaries
\usetikzlibrary{arrows.meta,positioning,fit,backgrounds}

\newcommand{\hl}[1]{#1}

\setcopyright{none}
\renewcommand\footnotetextcopyrightpermission[1]{}

\newacronym{cti}{CTI}{Cyber Threat Intelligence}
\newacronym{oscr}{OSC\&R}{Open Software Supply Chain Attack Reference}

\acmConference[TRUST '27]{ACM Conference on Trustworthy and Responsible AI and Computing Systems}{March 7--9, 2027}{Washington, DC, USA}

\begin{document}

\title[Sentence-Level CTI Classification Across ATT\&CK and OSC\&R]{Bridging MITRE ATT\&CK and OSC\&R: A Sentence-Level Cyber Threat Intelligence Classification Study}

\author{Nimmy Binte Islam}
\affiliation{%
  \institution{Premier University}
  \city{Chittagong}
  \country{Bangladesh}}
\email{islam.nimmybinte@gmail.com}

\author{Md Erfan}
\affiliation{%
  \department{Department of Computer Science}
  \institution{The University of Alabama}
  \city{Tuscaloosa}
  \state{Alabama}
  \country{USA}}
\email{merfan@crimson.ua.edu}

\author{Ahmed Ryan}
\affiliation{%
  \department{Department of Computer Science}
  \institution{The University of Alabama}
  \city{Tuscaloosa}
  \state{Alabama}
  \country{USA}}
\email{aryan9@crimson.ua.edu}

\author{Md Rayhanur Rahman}
\authornote{Corresponding author.}
\affiliation{%
  \department{Department of Computer Science}
  \institution{The University of Alabama}
  \city{Tuscaloosa}
  \state{Alabama}
  \country{USA}}
\email{mrahman87@ua.edu}

\begin{abstract}
\Gls{cti} reports describe attacker behavior in natural language, which analysts map to frameworks such as MITRE ATT\&CK. 
ATT\&CK covers a broad range of cyberattacks and describes supply chain compromise at a coarse level, whereas the \gls{oscr} provides a fine-grained classification focused on software supply chain attacks. We found no prior study that classifies CTI sentences across both frameworks. We address this gap by annotating sentence-level segments as ATTCK, OSCR, BOTH, or NON-ATTACK, assigning applicable technique identifiers, and evaluating models on these tasks. 
We built a corpus of 2,254 sentence-level segments from 20 CTI reports, held out 4 reports (451 segments) as a test set, and performed 4-fold cross-validation on the remaining 16 reports(1,803 segments), keeping segments from each report together. 
We compared a majority-class baseline, a TF-IDF linear SVM, three CTI-BERT models (single-head, multi-task, and two-stage), and a QLoRA-adapted Qwen2.5-7B-Instruct model. 
The two-stage CTI-BERT model achieved the highest average macro-F1 score on the held-out test set at 0.6392, whereas the TF-IDF model scored 0.4846 and the QLoRA model scored 0.1985.
For technique identification, the multi-task model achieved macro-F1 scores of 0.0146 (ATT\&CK) and 0.0174 (OSC\&R), and micro-F1 scores of 0.0215 and 0.0212, respectively. The multi-task model most often made errors by classifying BOTH segments as ATTCK only.
\end{abstract}



\maketitle
\section{Introduction}
\label{sec:intro}

Modern organizations rely on software systems for everyday operations, making them common targets for cyberattacks. To protect these systems, security analysts need to understand how attackers operate. \gls{cti} reports published by security vendors, national response teams, and independent researchers describe real-world attacks. Analysts use these reports to develop detection rules and track the activities of known threat groups \cite{alsada2024mitre}. However, \gls{cti} reports describe attacker actions in plain language, so analysts must identify these actions and assign each relevant sentence to a standard label indicating the attack behavior it describes. To do this, analysts use MITRE ATT\&CK \cite{strom2020mitre}, a publicly available knowledge base that organizes known attacker behavior into tactics and techniques. A tactic describes what an attacker aims to achieve, while a technique describes how that goal is achieved. For example, gaining initial access is a tactic, while phishing is a technique \cite{strom2020mitre, mitre_t1566}. Analysts can assign these labels by hand, but the work grows with the number of reports, so reports often go unlabeled. Therefore, such reports are harder to search, compare, and turn into detection rules \cite{husari2017ttpdrill, rani2023ttphunter}. Automatically labeling individual sentences could reduce this effort and make CTI reports easier to use.

\begin{figure*}[!t]
    \centering
    \includegraphics[width=\textwidth]{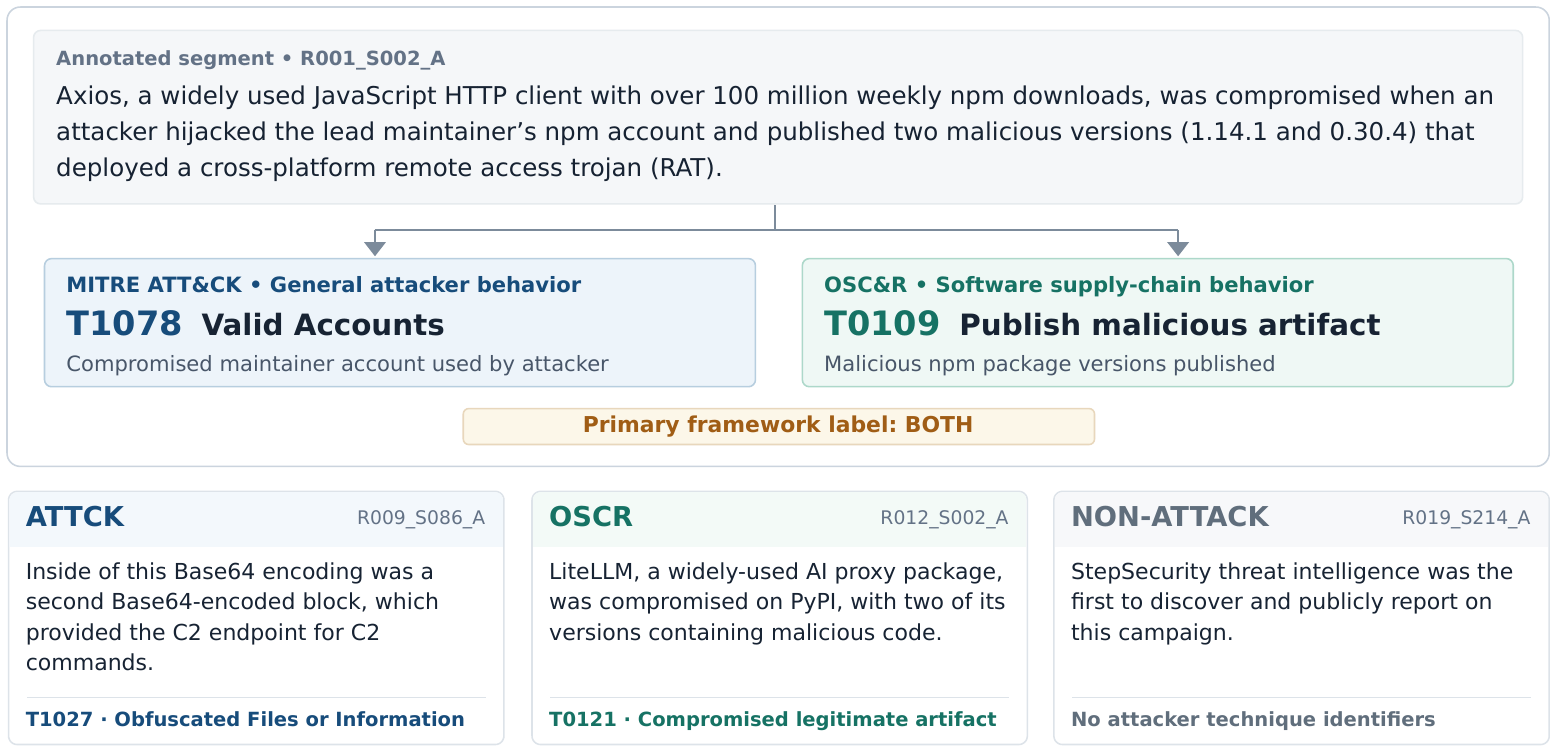}
    \caption{The top segment is labeled BOTH because it describes ATT\&CK (T1078) and OSC\&R (T0109) behavior. The bottom segments are labeled ATTCK, OSCR, and NON-ATTACK. The technique identifiers shown for the attack segments illustrate our secondary task of identifying applicable techniques within each framework.}
    \label{fig:motivating}
\end{figure*}

However, assigning the correct label becomes more difficult when these sentences describe software supply chain attacks. A software supply chain includes the code, dependencies, build systems, and update channels through which software reaches its users~\cite{haque2025autonomous}. Attackers can target different parts of this chain, such as build workflows \cite{koishybayev2022github} and third-party packages \cite{ohm2020backstabber, ryan2025unveiling, gao2025malguard}. A systematic study identified 107 distinct attack vectors across these stages \cite{ladisa2023sok}, showing how many ways a supply chain can be compromised. This variety creates a challenge for MITRE ATT\&CK, which covers a broad range of cyberattacks but provides limited detail about the different ways software supply chains can be compromised \cite{mitre_t1195}. For example, the SolarWinds attack involved a compromised software build process, while NotPetya spread through a malicious software update. Both are covered under the same broad supply-chain category in ATT\&CK, despite targeting different stages \cite{mitre_t1195_002}. The \gls{oscr} \cite{pbom_oscr} provides more detailed techniques for describing attacks across the software supply chain. While ATT\&CK covers general attacker behavior, OSC\&R focuses on how software supply chains are compromised. The two frameworks therefore complement each other. Figure~\ref{fig:motivating} shows a real CTI sentence from our corpus that needs labels from both frameworks, along with examples of the other three framework labels. Although both frameworks are needed to describe the different types of supply-chain attacks, existing automated CTI mapping methods mainly focus on ATT\&CK~\cite{tamanna2026review}, leaving OSC\&R largely unexplored.

Early studies on automated CTI mapping focused on classifying entire reports using MITRE ATT\&CK \cite{li2022attackg, ayoade2018automated, legoy2020automated}, while later research shifted toward identifying attack techniques in individual sentences \cite{rani2023ttphunter, li2024automated, huang2024mitretrieval, abdeen2024smet}. More recent studies have explored different approaches to improve this classification. These include comparing fine-tuned encoder models with large language models (LLMs) \cite{buechel2025sok} and generating synthetic sentences to improve the classification of rare techniques \cite{ruizrodenas2025synthcti}. Other studies have evaluated open-source LLMs on CTI sentences, including sentences with no identifiable attack technique \cite{ryan2026evaluating}, and examined technique extraction from reports covering the SolarWinds and XZ Utils attacks \cite{haque2026beyond}. LLMs have also been used to map vehicle vulnerability descriptions to structured threat information, including ATT\&CK techniques~\cite{erfan2026stix}. Despite these developments, the reviewed methods primarily focus on ATT\&CK, while OSC\&R remains available as framework documentation rather than an established classification benchmark \cite{pbom_oscr}. To the best of our knowledge, no previous study has jointly classified CTI sentences as ATT\&CK, OSC\&R, both, or neither, and identified the corresponding techniques within each framework.

The goal of our research is to support attack detection by helping security analysts and researchers determine whether each CTI report sentence describes ATT\&CK behavior, OSC\&R behavior, both, or no attack, and which techniques apply.

We are investigating the following research questions:

\smallskip

\textbf{RQ1: Dataset Construction and Framework Classification.} How can CTI report sentences be cleaned and labeled against both MITRE ATT\&CK and OSC\&R, and how well can different models classify them into the resulting four framework categories?

\textbf{RQ2: Technique Identification.} How accurately can models identify MITRE ATT\&CK and OSC\&R techniques from CTI sentences, particularly for frequently and rarely observed techniques?

\textbf{RQ3: Classification Challenges.} What are the key error patterns in sentence-level CTI classification, especially when distinguishing overlapping frameworks and identifying attack behaviors?

\smallskip

To address these research questions, we designed an experimental study covering dataset construction, model evaluation, technique identification, and error analysis.
We constructed a cleaned dataset of 2,254 sentence-level segments from 20 CTI reports. Each segment was assigned one of four framework labels (ATTCK, OSCR, BOTH, or NON-ATTACK), along with the corresponding technique identifiers where applicable. To prevent information leakage, we divided the dataset by report, reserving 4 reports containing 451 segments as a test set and using the remaining 16 reports for 4-fold cross-validation. For RQ1, we compared six models on the same four cross-validation folds, then trained each model on all 16 pool reports and evaluated it once on the reserved test set, using paired bootstrap confidence intervals and the exact McNemar test for statistical comparison. For RQ2, we evaluated technique identification performance for the multi-task and QLoRA models, examining results separately for frequently and rarely observed attack techniques. For RQ3, we analyzed the framework-level classification errors of the multi-task, two-stage, and QLoRA models to identify common error patterns and potential difficulties in distinguishing between different attack frameworks.

This study makes three main contributions. First, we introduce a joint sentence-level CTI classification task using both MITRE ATT\&CK and OSC\&R frameworks. This allows attack behavior to be classified into four categories (ATTCK, OSCR, BOTH, or NON-ATTACK), along with the identification of individual techniques within each framework for the same segments.

Second, we construct an annotated corpus of 2,254 segments, supported by a documented preprocessing pipeline and report-level data splitting to prevent information leakage between the evaluation partitions. This provides a foundation for evaluating cross-framework classification on previously unseen reports. The annotated dataset, together with the annotation instructions and guidelines, is available at \url{https://figshare.com/s/b6d0513d8d700663e6aa}.

Third, we compare six models on the same dataset and examine their performance through statistical comparisons, framework-level error analysis, and technique identification results for both frequent and rare techniques. These findings identify the relative strengths and limitations of different classification approaches and the key challenges that remain in distinguishing overlapping frameworks and identifying individual attack techniques.

This paper is organized as follows. Section~\ref{sec:related} defines key concepts and reviews relevant prior work on automated CTI mapping. Section~\ref{sec:method} presents the formulation, corpus, splitting protocol, models, selection procedure, and evaluation metrics. Section~\ref{sec:results} reports cross-validation and test set results, statistical comparisons, and error analysis. Section~\ref{sec:discussion} discusses the findings, Section~\ref{sec:threats} covers threats to validity, and Section~\ref{sec:conclusion} concludes.

\section{Background and Related Work}
\label{sec:related}

This section introduces the key concepts of our study and reviews research on automated CTI classification and technique mapping.

\subsection{Key Concepts}
\label{subsec:keyconcepts}

This section explains the main CTI terms, the two frameworks, and the classification tasks we use throughout.

\vspace{0.5\baselineskip}

\hl{\mbox{\emph{CTI reports and TTPs.}} CTI reports describe adversary activity and are published by security vendors, national response teams, and independent researchers. The reports used in this study are grey literature, that is, publicly accessible technical reports produced by commercial security vendors rather than peer-reviewed articles. Their behavioral content is expressed as tactics, techniques, and procedures (TTPs), where a procedure is a specific observed implementation of a particular technique.}

\hl{\mbox{\emph{Frameworks and technique identifiers.}} MITRE ATT\&CK is a knowledge base that organizes adversary behavior into tactics, techniques, and sub-techniques derived from real-world observations~\cite{strom2020mitre}. This study uses its enterprise matrix, in which software supply chain compromise is represented mainly by the Supply Chain Compromise technique T1195 and its sub-techniques~\cite{mitre_t1195}. OSC\&R is an ATT\&CK-like matrix scoped to the software supply chain, covering build and deployment systems, third-party components, and malicious or compromised updates~\cite{pbom_oscr}. Both frameworks use technique identifiers starting with T followed by four digits. In our corpus, ATT\&CK identifiers follow the T1nnn pattern, while OSC\&R identifiers follow T0nnn. We remove these technique identifiers from the segment text (Section~\ref{subsec:corpus}) so models cannot use them to distinguish between the frameworks.}

\hl{\mbox{\emph{Tasks and segments.}} Framework classification, the primary task, assigns each segment one of four labels, ATTCK, OSCR, BOTH, or NON-ATTACK, according to which framework, if any, its attack behavior belongs to. Technique identification, the secondary task, predicts the specific technique identifiers within each framework, and a segment can have more than one, since attacks often combine several techniques~\cite{erfan2026cooccur}. A segment is usually one sentence, but 52 long sentences that describe multiple attacker actions were split into smaller segments (Section~\ref{subsec:corpus}).}

\subsection{Related Work}
\label{subsec:relatedwork}

We organize the relevant prior work around our three research questions. Table~\ref{tab:related-work} summarizes the studies discussed in this section and the corresponding scope of our study.

\begin{table}[t]
  \caption{Prior Work on Automated CTI Mapping and the Corresponding Scope of This Study}
  \label{tab:related-work}
  \begin{tabular}{@{}>{\raggedright\arraybackslash}p{0.35\columnwidth}
                  >{\raggedright\arraybackslash}p{0.17\columnwidth}
                  >{\raggedright\arraybackslash}p{0.37\columnwidth}@{}}
    \toprule
    \textbf{Prior work} & \textbf{Supported by} & \textbf{This study} \\
    \midrule
        Annotated CTI datasets label attack techniques against ATT\&CK only
      & \cite{alam2024ctibench,dellapenna2025ctihal,lange2024annoctr,mitre_tram,ryan2026evaluating}
      & Labels each segment against ATT\&CK and OSC\&R as ATTCK, OSCR, BOTH, or NON-ATTACK \\
    \addlinespace
    Encoder and LLM classifiers are compared on ATT\&CK labels only
      & \cite{buechel2025sok}
      & Compares classical, encoder, and LLM models on the four-way decision \\
    \addlinespace
    Technique labels are sparse and imbalanced
      & \cite{buechel2025sok,huang2024mitretrieval,legoy2020automated,li2024automated,ryan2026evaluating}
      & Reports frequent and rare techniques separately for both frameworks \\
    \addlinespace
        Error analyses examine confusion within ATT\&CK
      & \cite{buechel2025sok,haque2026beyond,huang2024mitretrieval,li2024automated,ryan2026evaluating}
      & Analyzes confusion between ATT\&CK and OSC\&R \\
    \bottomrule
  \end{tabular}
\end{table}

\paragraph{Dataset Construction and Framework Classification.} Public CTI datasets label attack techniques against ATT\&CK only. TRAM~\cite{mitre_tram} covers up to 50 common techniques, AnnoCTR~\cite{lange2024annoctr} links entities, tactics, and techniques in 120 of its 400 reports to ATT\&CK, CTI-HAL~\cite{dellapenna2025ctihal} links sentences that reference TTPs in 81 reports to techniques, and CTIBench~\cite{alam2024ctibench} includes a technique extraction task over 60 malware descriptions. Ryan et al.~\cite{ryan2026evaluating} labeled 2,076 sentences from 83 reports, including 795 that describe no technique. We found no sentence-level CTI dataset labeled against OSC\&R. The closest study, Wang et al.~\cite{wang2023portrait}, used ChatGPT and expert review to identify attack phases and techniques from 1,520 incident texts covering 277 software supply chain attacks. They developed a matrix of 14 tactics and 113 techniques drawing on ATT\&CK and OSC\&R. However, their study analyzed entire incident texts rather than individual sentences and did not train a classifier. Our corpus labels each sentence against both ATT\&CK and OSC\&R, with separate labels for sentences that fit both frameworks (BOTH) and sentences that fit neither (NON-ATTACK). Because a report label does not show which passage supports a technique, classification methods have moved from whole reports~\cite{ayoade2018automated, legoy2020automated, li2022attackg} to individual threat actions~\cite{husari2017ttpdrill, zhang2021exaction} and sentences~\cite{rani2023ttphunter, li2024automated, huang2024mitretrieval, abdeen2024smet}. We adopt sentence granularity from this work. The new element of our corpus is the dual-framework label, not the sentence unit. Three model families are relevant: count-based features with classical classifiers~\cite{kim2022comparative}, security-specific encoders such as CyBERT~\cite{ranade2021cybert}, SecureBERT~\cite{aghaei2023securebert}, and CTI-BERT~\cite{park2023ctibert}, and instruction-tuned LLMs such as Qwen2.5~\cite{yang2024qwen25} adapted through low-rank adaptation~\cite{hu2022lora} or QLoRA~\cite{dettmers2023qlora}. B{\"u}chel et al.~\cite{buechel2025sok} found that fine-tuned LLMs perform comparably to BERT-style classifiers but are harder to build and run. This comparison uses ATT\&CK labels only, whereas in our study we compare all three families on a four-way classification decision across ATT\&CK and OSC\&R.

\paragraph{Technique Identification.} Annotated data are scarce and technique labels are imbalanced across this literature. Legoy et al.~\cite{legoy2020automated} found techniques that appear in only one report and removed techniques with fewer than five reports, and earlier studies limited their scope to 5, 10, or 50 techniques, as in TTPHunter~\cite{rani2023ttphunter}, because adequately labeled data were scarce~\cite{huang2024mitretrieval}. Li et al.~\cite{li2024automated} attribute weaker technique than tactic classification to the much larger technique label space, in which sparser classes have insufficient training examples. In the corpus of Ryan et al.~\cite{ryan2026evaluating}, 56 of 114 techniques have five or fewer instances. B{\"u}chel et al.~\cite{buechel2025sok} identify the lack of high-quality annotated data as the main bottleneck and report precision well below 40\% for less common techniques in realistic settings. Proposed remedies include oversampling through text augmentation~\cite{kim2022comparative}, synthetic sentences for underrepresented techniques, which raised the macro-F1 of a SecureBERT classifier from 0.44 to 0.66~\cite{ruizrodenas2025synthcti}, and fusion of a sentence-level BERT model with ontology knowledge~\cite{huang2024mitretrieval}. We found no existing trained technique classifier for OSC\&R, and our technique heads use none of these proposed remedies. Both of our vocabularies are sparse, so we report results separately for frequent and rare techniques.

\paragraph{Classification Challenges.} Techniques with overlapping descriptions are hard to separate. MITREtrieval~\cite{huang2024mitretrieval} notes that T1132 (Data Encoding) and T1027 (Obfuscated Files or Information) can share the verb-object pair, which makes term-frequency features such as those of TTPDrill~\cite{husari2017ttpdrill} ambiguous. Li et al.~\cite{li2024automated} traced sentences misclassified as Masquerading (T1036) to overlap in the scope of techniques such as Hide Artifacts (T1564) and Protocol Impersonation (T1001.003). Haque et al.~\cite{haque2026beyond} found that up to 33.3\% of misclassifications involved techniques that share tactics and have overlapping descriptions. Ryan et al.~\cite{ryan2026evaluating} grouped LLM errors into false positives, in which models misread keywords or assign techniques to benign descriptions, and false negatives, in which they miss described techniques. B{\"u}chel et al.~\cite{buechel2025sok} report that frequently confused pairs, such as T1547.001 and T1112, which both modify the Windows Registry, cannot be separated from a single sentence even by an expert, and they suspect that ATT\&CK is not always sufficiently unambiguous. These error analyses examine confusion within ATT\&CK, whereas we analyze confusion between ATT\&CK and OSC\&R, together with false alarms and missed attacks.

\section{Methodology}
\label{sec:method}
We describe the methodology for RQ1--RQ3 in this section.

\subsection{RQ1: Dataset Construction and Framework Classification}
\label{subsec:rq1}

\subsubsection{Dataset Construction}
\label{subsec:corpus}

We construct it through six steps.

\noindent\textbf{Step 1: Search Query Construction.}
We used Google Search to identify CTI reports published between 2023 and 2026. General search queries were refined to software supply chain incidents by combining terms such as npm, PyPI, GitHub Actions, CI/CD, and malicious packages with compromise-related terms such as malware, backdoor, and threat actor. Searches were restricted by date or vendor website, and later included specific incidents or known threat actors, such as XZ Utils and TeamPCP.

\noindent\textbf{Step 2: Report Collection and Screening.}
We collected 203 candidate records, retaining 199 after removing duplicate URLs. Initial screening required publicly available English-language reports describing actual or technically analyzed software supply chain attacks and attacker actions. We excluded opinion, awareness, product, and mitigation-only articles, vulnerability disclosures without executed compromise, and unrelated malware reports, retaining 104 records. The second screening selected primary technical analyses describing both the supply chain compromise and subsequent attacker behavior. For repeated incidents, we retained the most detailed report unless different campaign stages were covered. The final 20 reports, published by six security vendors, cover package registries, source-hosting and CI/CD platforms, build processes, software updates, and extension marketplaces. Each report contained at least five segments ultimately labeled BOTH.

\noindent\textbf{Step 3: Preprocessing.}
We extracted HTML content, removed boilerplate, and retained relevant paragraphs, lists, and tables. Indicators of compromise were replaced with typed placeholders. Using spaCy, we segmented the text into sentences, removed sentences shorter than five words, and further split long sentences describing multiple distinct attacker actions into separately labelable segments. This produced 2,255 segments from 2,198 sentences, including 52 sentences requiring further splitting.  We then applied a rule-based filter, which uses fixed text patterns to identify segments that are unlikely to describe attacker behavior. Such low-signal text includes recommendations, report boilerplate, patch or version notices, and attribution or contact statements, for example, a sentence advising readers to upgrade an affected package. The filter pre-assigned NON-ATTACK to segments matching these patterns and to short background statements that mention no threat actor, indicator placeholder, or attack-related keyword. All other segments were sent for manual review, with keyword lists marking whether each one might relate to ATT\&CK, OSC\&R, or both.

\noindent\textbf{Step 4: Annotation.}
The primary annotator independently labeled all segments into the four classes (ATTCK, OSCR, BOTH, NON-ATTACK). Defender activities were labeled NON-ATTACK. Supply chain mechanisms were assigned OSC\&R techniques, reserving ATT\&CK T1195 for compromises without a specified mechanism. Every NON-ATTACK label pre-assigned by the rule-based filter was manually checked. We refer to the primary annotator's framework label and technique identifiers for each segment as its gold label and gold techniques. They serve as the reference that models are trained on and evaluated against. For example, the top segment in Figure~\ref{fig:motivating} has the gold label BOTH, with gold techniques including T1078 for ATT\&CK and T0109 for OSC\&R, so a model that predicts ATTCK for this segment is counted as wrong.

\noindent\textbf{Step 5: Inter-Annotator Agreement.}
The primary annotator independently annotated 2,255 segments from 20 reports. After the initial annotation, a second annotator independently labeled a reproducible random sample of 225 segments (10\%), drawn from 20 reports, without access to the primary labels. Framework agreement was measured using Cohen's $\kappa$, while technique agreement was calculated as the unweighted mean of per-technique binary $\kappa$ values. Framework agreement reached 92.0\% ($\kappa = 0.857$), while the mean technique-level agreement was $\kappa = 0.654$, comprising 0.670 across 31 ATT\&CK techniques and 0.638 across 30 OSC\&R techniques.

\noindent\textbf{Step 6: Cleaning and Validation.}
We normalized framework labels (whitespace and case) and technique identifiers (delimiter splitting, case folding), collapsed sub-techniques to their parent identifiers, and removed 39 identifier tokens from 30 segments to prevent label leakage. We removed one blank row, leaving a final dataset of 2,254 segments from 20 reports.

\subsubsection{Report-Level Splitting}
\label{subsec:splitting}
Segments from the same report share incident details, vocabulary, and writing style, so random sentence-level splitting would place near-duplicate segments on both sides of the train/test boundary and inflate apparent performance. We therefore split at the report level. A test set of 4 reports was selected once, before any training, as the combination closest to a 20\% test fraction among those that cover all four framework labels and include at least one ATTCK-majority and one OSCR/BOTH-majority report. The test set contains 451 of the 2,254 segments (20.01\%). The remaining 16 reports (1,803 segments) form the cross-validation pool, divided into four report-grouped folds~\cite{pedregosa2011scikit} with 456, 447, 454, and 446 segments from 3, 3, 5, and 5 reports. Every partition contains all four labels (Table~\ref{tab:coverage}).

\subsubsection{Classification Models}
\label{subsec:models}
We trained and tested six models on the four-way framework label (ATTCK, OSCR, BOTH, NON-ATTACK) and then describe their training procedure.

\paragraph{Reference baselines.} The majority-class baseline predicts the most common training label for every segment and serves as a basic point of comparison for the learned models. The TF-IDF + LinearSVC model is a linear support vector classifier with balanced class weights over word unigram and bigram TF-IDF features (at most 5,000)~\cite{pedregosa2011scikit} and serves as the reference model.

\paragraph{Single-label CTI-BERT models.} The single-head model fine-tunes CTI-BERT~\cite{park2023ctibert} with one four-way output layer and class-weighted cross-entropy, where each class weight is inversely proportional to the class frequency in the training partition~\cite{pedregosa2011scikit} and is recomputed for every fold and for the final retrain. The checkpoint with the highest development macro-F1 (Section~\ref{subsec:eval}) is restored. The two-stage cascade uses two CTI-BERT classifiers trained with the same class-weighted loss. Stage one separates NON-ATTACK from ATTACK; stage two assigns ATTCK, OSCR, or BOTH and is trained only on segments whose gold label is an attack label.

\paragraph{Multi-task model.} A shared CTI-BERT encoder feeds four linear heads from the [CLS] embedding, namely one presence head per framework and one multi-label technique head per framework (47 ATT\&CK and 41 OSC\&R outputs, see Section~\ref{subsec:rq2}). Each presence head is trained with binary cross-entropy in which positive examples are weighted by the negative-to-positive ratio of that framework's presence label in the training partition. The four losses are summed with equal weights during cross-validation and with the weights selected in Section~\ref{subsec:hpsearch} in the final model. The framework label is derived from the two presence probabilities. It is BOTH if both exceed their thresholds, ATTCK or OSCR if only one does, and NON-ATTACK otherwise. In cross-validation, the two thresholds are jointly grid-searched from 0.10 to 0.90 in steps of 0.05 to maximize development macro-F1 (Section~\ref{subsec:hpsearch}).

\paragraph{QLoRA-adapted LLM.} Qwen2.5-7B-Instruct~\cite{yang2024qwen25} is loaded in 4-bit NF4 precision with double quantization~\cite{dettmers2023qlora}, and a LoRA adapter~\cite{hu2022lora} of rank 16, scaling factor 32, and dropout 0.1 is attached to all attention and MLP projection matrices. Only the adapter is trained. Given a segment wrapped in a fixed system prompt, the model is trained to generate a JSON object containing the framework label and the two technique lists, with the loss computed only on the generated tokens. Decoding is greedy with a 128-token limit. An output from which no valid JSON object can be parsed is scored as incorrect, matching no framework label and containing no technique identifiers. The model uses the same settings in cross-validation and in the final retrain, namely a maximum length of 256, paged 8-bit AdamW with learning rate $2\times10^{-4}$, batch size 1 with gradient accumulation over 8 steps, one epoch, and seed 42, with no development-based selection.

\paragraph{Training procedure.} For each model described below, we trained and evaluated it across four cross-validation folds, then trained a final model on all 16 reports for evaluation on the test set. The three encoders use AdamW with learning rate $2\times10^{-5}$, batch size 16, weight decay 0.01, 10\% linear warmup, and mixed precision during model training. In cross-validation, they use a maximum length of 128 tokens and seed 42, and train for up to 100 epochs with early stopping (patience 10 on macro-F1). 

\subsubsection{Hyperparameter Search and Model Selection}
\label{subsec:hpsearch}
With 16 reports in the cross-validation pool, the search was kept small and run on only a single development fold, without access to the test reports or the other three folds. Twelve random-search trials tuned the multi-task model's maximum length (128 or 256), learning rate (log-uniform in [$1\times10^{-5}$, $5\times10^{-5}$]), batch size (8, 16, or 32), dropout (0.1 or 0.3), and the presence and technique loss weights (each 0.5, 1.0, 1.5, or 2.0), training up to 10 epochs with patience 2. The best trial (maximum length 256, learning rate $2.72\times10^{-5}$, batch size 16, dropout 0.3, loss weights 1.0 and 0.5) reached a development macro-F1 score of 0.6311. Thresholds tuned on its development predictions were 0.85 (ATT\&CK presence), 0.90 (OSC\&R presence), and 0.10 for both technique heads. The selected configuration was fixed before any test-set evaluation.

\subsubsection{Evaluation Metrics and Statistical Tests}
\label{subsec:eval} 
Because the framework labels are imbalanced, accuracy is not informative on its own~\cite{saito2015precision}, so macro-F1 over the four labels is the primary metric. Macro-F1 is the unweighted mean of the per-label F1 scores, where F1 combines precision and recall, so each label counts equally regardless of its size. We also report weighted F1, accuracy, and the Matthews correlation coefficient (MCC). Cross-validation results are the mean and standard deviation over the four development folds, and each model is evaluated once on the test set. Test-set comparisons between two models use two paired procedures on the 451 segments~\cite{dietterich1998approximate}. A bootstrap draws 2,000 resamples of the test segments with a fixed seed and computes the macro-F1 difference in each. We report the mean difference ($\Delta$) and the 2.5th--97.5th percentile interval. An exact McNemar test compares the number of segments that only model A labels correctly with the number that only model B labels correctly. We use the exact two-sided binomial form because the smallest pair has only 13 such segments (Table~\ref{tab:significance}).

\subsection{RQ2: Technique Identification}
\label{subsec:rq2}
We evaluate technique identification to address RQ2.

\paragraph{Technique label space.} Technique identifiers are collapsed to their parent technique (Section~\ref{subsec:corpus}, Step 6). Each framework's vocabulary is built from the 16-report training pool and contains 47 ATT\&CK and 41 OSC\&R techniques.

\paragraph{Technique prediction.} In the multi-task model, each technique head outputs one probability per technique in its framework's vocabulary. A head receives loss only on training segments whose gold label includes its framework, so the OSC\&R head, for example, is not trained on ATTCK-only segments. A technique is predicted when its probability exceeds a per-head threshold, grid-searched over the same 0.10--0.90 range to maximize development micro-F1 on the segments whose gold label includes that framework. The QLoRA model generates both technique lists directly as part of its final structured JSON output.

\paragraph{Technique metrics.} On the test set, technique prediction is evaluated as a multi-label task over all test segments, separately for each framework. Because most techniques have few examples, the primary metric is micro-F1, which pools true positives, false positives, and false negatives across techniques. We report macro-F1, which weights every technique equally. To compare frequent and rare techniques (occurs in five or fewer training segments), each vocabulary is split by training-pool segments. This follows from the evaluation design. The data are divided into five partitions, the four cross-validation folds and the held-out test set, so a technique appears in more than one segment per partition on average only if it appears in more than five segments in total. ATT\&CK technique micro-F1 of the multi-task and QLoRA models is compared with the paired bootstrap of Section~\ref{subsec:eval}.

\subsection{RQ3: Classification Challenges}
\label{subsec:rq3}
We analyze framework-level errors of the multi-task, two-stage, and QLoRA models on the test set to address RQ3.

\paragraph{Error taxonomy.} Each error is assigned to one of three types: a false alarm (gold NON-ATTACK, predicted an attack label), a missed attack (gold attack label, predicted NON-ATTACK), or a framework confusion (gold and predicted labels are different attack labels). An unparseable QLoRA output counts as a false alarm when the gold label is NON-ATTACK and as a framework confusion otherwise.

\paragraph{Error slices.} For the multi-task model, we compute error rates by gold label, sentence-length tercile, and technique rarity, where a segment's rarity is the lowest training-pool support among its gold technique identifiers, using the threshold of Section~\ref{subsec:rq2}. For all three models, we list frequent gold-to-predicted confusion pairs.

\section{Results}
\label{sec:results}
We report the findings for RQ1--RQ3 in turn.

\subsection{Findings of RQ1: Dataset Construction and Framework Classification}
\label{subsec:res-rq1}

\subsubsection{Dataset Characteristics}
\label{subsec:dataset-stats}
Table~\ref{tab:coverage} reports the framework label distribution of the 2,254 segments and its division across partitions. The corpus is imbalanced: NON-ATTACK accounts for 62.95\% of segments, followed by BOTH (16.73\%), ATTCK (13.13\%), and OSCR (7.19\%). At the technique level, 50 distinct ATT\&CK and 41 distinct OSC\&R techniques occur in the corpus, of which 27 (54\%) and 15 (37\%) occur in five or fewer segments of the full corpus.

\begin{table}[!h]
\centering
\caption{Framework Label Distribution by Partition}
\label{tab:coverage}
\fontsize{8}{9.5}\selectfont
\setlength{\tabcolsep}{3.5pt}
\renewcommand{\arraystretch}{1.03}
\begin{tabularx}{\columnwidth}{|X|c|c|c|c|c|}\hline
\textbf{Partition} & \textbf{ATTCK} & \textbf{OSCR} & \textbf{BOTH} & \textbf{NON-ATT.} & \textbf{Total} \\ \hline
Full corpus & 296 & 162 & 377 & 1419 & 2254 \\ \hline
Test set & 63 & 44 & 74 & 270 & 451 \\ \hline
Fold 0 dev & 109 & 25 & 48 & 274 & 456 \\ \hline
Fold 1 dev & 28 & 41 & 70 & 308 & 447 \\ \hline
Fold 2 dev & 52 & 24 & 83 & 295 & 454 \\ \hline
Fold 3 dev & 44 & 28 & 102 & 272 & 446 \\ \hline
\end{tabularx}

\vspace{2pt}
\begin{minipage}{\columnwidth}\fontsize{8}{9.5}\selectfont
ATTCK = ATT\&CK behavior only; OSCR = OSC\&R behavior only; BOTH = behavior from both frameworks; NON-ATT. = no attack behavior; Fold $k$ dev = development set of cross-validation fold $k$. Values are segment counts.
\end{minipage}
\end{table}

\subsubsection{Cross-Validation Performance}
\label{subsec:cv}
Table~\ref{tab:cv} reports framework macro-F1 over the four cross-validation folds. All three CTI-BERT models exceeded the majority-class baseline and the TF-IDF reference on every fold, and the multi-task model had the highest mean macro-F1 (0.5877 $\pm$ 0.0367). The QLoRA model exceeded the baseline but fell below the TF-IDF reference on every fold.

\begin{table}[!h]
\centering
\caption{Four-Fold Grouped-CV Framework Macro-F1}
\label{tab:cv}
\fontsize{8}{9.5}\selectfont
\setlength{\tabcolsep}{3pt}
\renewcommand{\arraystretch}{1.03}
\begin{tabularx}{\columnwidth}{|X|c|c|c|c|c|}\hline
\textbf{Model} & \textbf{F0} & \textbf{F1} & \textbf{F2} & \textbf{F3} & \textbf{Mean $\pm$ SD} \\ \hline
Majority class & 0.188 & 0.204 & 0.197 & 0.189 & 0.1945 $\pm$ 0.0065 \\ \hline
TF-IDF + SVC & 0.436 & 0.432 & 0.463 & 0.466 & 0.4493 $\pm$ 0.0153 \\ \hline
Single-head & 0.593 & 0.538 & 0.531 & 0.541 & 0.5507 $\pm$ 0.0245 \\ \hline
Multi-task & 0.639 & 0.590 & 0.535 & 0.586 & 0.5877 $\pm$ 0.0367 \\ \hline
Two-stage & 0.623 & 0.555 & 0.526 & 0.550 & 0.5635 $\pm$ 0.0360 \\ \hline
QLoRA & 0.324 & 0.371 & 0.319 & 0.390 & 0.3510 $\pm$ 0.0304 \\ \hline
\end{tabularx}

\vspace{2pt}
\begin{minipage}{\columnwidth}\fontsize{8}{9.5}\selectfont
F0 to F3 = cross validation folds; Macro-F1 = unweighted mean of per-label F1 scores across four framework classes. Mean $\pm$ SD = mean and population standard deviation over four folds; Majority class = predicts the most frequent training label for every segment; TF-IDF + SVC = TF-IDF features with a linear support vector classifier; Single-head, Multi-task, and Two-stage are CTI-BERT models; QLoRA = QLoRA-adapted Qwen2.5
\end{minipage}
\end{table}

\subsubsection{Test-Set Performance}
\label{subsec:test}
Table~\ref{tab:test} reports framework classification metrics on the 451-segment test set. The two-stage cascade had the highest point estimates on all four reported metrics, with macro-F1 0.6392, followed by the single-head CTI-BERT model(0.6244) and the multi-task model (0.6065). The QLoRA model reached 0.1985, close to the majority-class baseline (0.1872), and a total ok 29 of its 451 outputs (6.4\%) could not be parsed.

\begin{table}[!h]
\centering
\caption{Test-Set Framework Metrics}
\label{tab:test}
\fontsize{8}{9.5}\selectfont
\setlength{\tabcolsep}{5pt}
\renewcommand{\arraystretch}{1.03}
\begin{tabularx}{\columnwidth}{|X|c|c|c|c|}\hline
\textbf{Model} & \textbf{Macro-F1} & \textbf{Wtd-F1} & \textbf{Acc.} & \textbf{MCC} \\ \hline
Majority class & 0.1872 & 0.4484 & 0.5987 & 0.0000 \\ \hline
TF-IDF + SVC & 0.4846 & 0.6328 & 0.6475 & 0.3696 \\ \hline
Single-head & 0.6244 & 0.7444 & 0.7428 & 0.5638 \\ \hline
Multi-task & 0.6065 & 0.7364 & 0.7561 & 0.5652 \\ \hline
Two-stage & 0.6392 & 0.7557 & 0.7583 & 0.5823 \\ \hline
QLoRA & 0.1985 & 0.4572 & 0.5743 & 0.0847 \\ \hline
\end{tabularx}

\vspace{2pt}
\begin{minipage}{\columnwidth}\fontsize{8}{9.5}\selectfont
Macro-F1 = unweighted mean of the per-label F1 scores; Wtd-F1 = per-label F1 weighted by the number of test segments with each gold label; Acc. = Accuracy; MCC = Matthews correlation coefficient.
\end{minipage}
\end{table}

\subsubsection{Pairwise Model Comparison}
\label{subsec:stats}
Table~\ref{tab:significance} compares model pairs on the same 451 test segments with the paired bootstrap and the exact McNemar test of Section~\ref{subsec:eval}. The multi-task model and the cascade both outperform the QLoRA model, with confidence intervals excluding zero and p $<$ 0.01, and the single-head CTI-BERT model outperforms the TF-IDF reference. The three CTI-BERT models do not differ detectably, as every pairwise interval among them spans zero, and every McNemar p-value exceeds 0.3. The two tests differ for the QLoRA model against the majority-class baseline. The macro-F1 interval includes zero, but on the 13 segments where only one of the two is correct, the majority-class baseline is correct on 12 of these segments and the QLoRA model on 1 (exact McNemar p = 0.0034).

\begin{table}[!h]
\centering
\caption{Pairwise Comparisons on the Test Set}
\label{tab:significance}
\fontsize{8}{9.5}\selectfont
\setlength{\tabcolsep}{2.5pt}
\renewcommand{\arraystretch}{1.03}
\begin{tabularx}{\columnwidth}{|>{\raggedright\arraybackslash}X|c|c|c|c|}\hline
\textbf{Pair (A vs. B)} & \textbf{$\Delta$F1} & \textbf{95\% CI} & \textbf{b, c} & \textbf{p} \\ \hline
Multi-task vs. two-stage & $-0.0338$ & $[-0.090, 0.021]$ & 29, 30 & 1.000 \\ \hline
Multi-task vs. QLoRA & $+0.4055$ & $[0.347, 0.464]$ & 91, 9 & $3.3{\times}10^{-18}$ \\ \hline
Two-stage vs. QLoRA & $+0.4394$ & $[0.383, 0.497]$ & 105, 22 & $3.7{\times}10^{-14}$ \\ \hline
Single-head vs. TF-IDF & $+0.1405$ & $[0.066, 0.214]$ & 84, 41 & $1.5{\times}10^{-4}$ \\ \hline
Multi-task vs. single-head & $-0.0190$ & $[-0.077, 0.042]$ & 39, 33 & 0.556 \\ \hline
Two-stage vs. single-head & $+0.0148$ & $[-0.029, 0.061]$ & 26, 19 & 0.371 \\ \hline
QLoRA vs. majority class & $+0.0109$ & $[-0.001, 0.032]$ & 1, 12 & 0.0034 \\ \hline
\end{tabularx}

\vspace{2pt}
\begin{minipage}{\columnwidth}\fontsize{8}{9.5}\selectfont
$\Delta$F1 = macro-F1 of model A (first model in a row) minus model B (second model in a row); CI = Confidence interval; b = segments that only model A labels correctly. c = segments that only model B labels correctly. p = two-sided exact McNemar $p$-value computed from b and c.
\end{minipage}
\end{table}

\subsection{Findings of RQ2: Technique Identification}
\label{subsec:res-rq2}

\paragraph{Overall, frequent, and rare techniques.} On the test set, the multi-task model reached technique micro-F1 of 0.0215 for ATT\&CK and 0.0212 for OSC\&R (macro-F1 0.0146 and 0.0174). Under the rarity threshold of Section~\ref{subsec:rq2}, 26 of the 47 ATT\&CK and 16 of the 41 OSC\&R training-pool techniques are rare. On frequent techniques, the multi-task model reached micro-F1 of 0.0415 (ATT\&CK) and 0.0319 (OSC\&R), and on rare techniques 0.0006 and 0.0004. The QLoRA model produced no correct technique identifiers in either framework, so every QLoRA technique score is 0.0000.

\paragraph{Cross-validation versus test.} The QLoRA model scored differently on the four development folds and the test set. In cross-validation, the model reached mean technique micro-F1 score of 0.2507 $\pm$ 0.0881 for ATT\&CK and 0.0194 $\pm$ 0.0242 for OSC\&R, but 0.0000 for both frameworks on the test set.

\paragraph{Paired comparison.} The paired bootstrap of Section~\ref{subsec:stats}, applied to ATT\&CK technique micro-F1, gives $\Delta = +0.0216$ (95\% CI [0.018, 0.026]) for the multi-task model over the QLoRA model. The interval excludes zero, but the observed difference is relatively small because the multi-task score is itself 0.0215.

\subsection{Findings of RQ3: Classification Challenges}
\label{subsec:erroranalysis}

\paragraph{Error types.} On the test set, the multi-task model made 11 false alarms, 47 missed attacks, and 52 framework confusions; the cascade made 24, 31, and 54; and the QLoRA model made 12, 153, and 27. Framework confusion is the largest error type for both CTI-BERT models (52 of 110 errors for the multi-task model and 54 of 109 for the cascade). Missed attacks make up most QLoRA errors (153 of 192), as the model predicts NON-ATTACK for 411 of the 451 segments and never predicts ATTCK or BOTH. Among the QLoRA errors, 11 false alarms and 18 framework confusions come from unparseable outputs rather than predicted labels.

\paragraph{Error rates by label, length, and rarity.} For the multi-task model, error rates by gold label are BOTH 52/74 (70.3\%), OSCR 22/44 (50.0\%), ATTCK 25/63 (39.7\%), and NON-ATTACK 11/270 (4.1\%), so errors concentrate in attack labels. Error rates vary less with sentence length, at 29.1\% (41/141) for medium, 23.1\% (33/143) for long, and 21.6\% (36/167) for short segments. They vary little with technique rarity, at 57.1\% (12/21) for segments whose rarest technique is rare versus 54.4\% (87/160) for segments with only frequent techniques.

\paragraph{Most frequent confusion pairs.} The most frequent confusion is BOTH$\rightarrow$ATTCK for the multi-task model (30) and cascade (23), and BOTH$\rightarrow$NON-ATTACK for the QLoRA model (61). For every model, the leading confusion starts from a gold BOTH segment.

\section{Discussion}
\label{sec:discussion}
Our results answer RQ1--RQ3 through four implications.

\paragraph{Implication 1. Fine-tuned CTI-BERT models handle the framework decision, and the QLoRA model did not}
All three CTI-BERT models exceeded the TF-IDF reference on every fold and reached test macro-F1 between 0.6065 and 0.6392. The QLoRA model reached 0.1985 on the test set, and the model's accuracy (0.5743) fell below the 0.5987 obtained by predicting NON-ATTACK for every segment. The evidence favors CTI-BERT for the framework decision.

\paragraph{Implication 2. The choice among CTI-BERT designs remains open}
The multi-task model ranked highest in cross-validation (0.5877) and the cascade ranked highest on the test set (0.6392), but no pairwise test among the three CTI-BERT models found a significant difference. Both gaps fall within the fold-to-fold standard deviations of Table~\ref{tab:cv}, so the reversal is consistent with sampling variation. A test set of 451 segments from four reports cannot resolve differences of about 0.03, and the absence of a detected difference does not show that the designs are equivalent.

\paragraph{Implication 3. Technique identification needs more labeled data}
Technique micro-F1 stayed near 0.02 for the multi-task model and near zero on rare techniques. The corpus itself explains part of the result. Of the training-pool techniques, 26 of 47 for ATT\&CK and 16 of 41 for OSC\&R occur in five or fewer segments.

\paragraph{Implication 4. Most framework confusions involve segments that fit both frameworks}
Framework confusion was the largest error type for both CTI-BERT models, the multi-task model misclassified 70.3\% of BOTH segments, and BOTH$\rightarrow$ATTCK was the most frequent confusion for both models. A plausible reason is that supply chain intrusions often include general post-compromise behavior that ATT\&CK also describes, so the OSC\&R evidence in a BOTH sentence can be limited. For the multi-task model, the decoding rule also predicts BOTH only when both presence scores exceed high thresholds (0.85 and 0.90). The cascade uses no thresholds and shows the same leading confusion, so the thresholds add to the confusion without being the only cause. Prior error analyses also report confusion among overlapping ATT\&CK techniques~\cite{li2024automated, haque2026beyond}, and our results show a similar overlap between the two frameworks.

\section{Threats to Validity}
\label{sec:threats}
This section discusses how far our findings generalize beyond the studied reports and the limitations of our evaluation methods.
\paragraph {External validity}
The corpus contains 20 grey-literature reports from 2023 to 2026 on software supply chain incidents, and the test set contains 4 reports. The results may not transfer to government advisories, incident-response reports, or other attack domains. OSC\&R has no prior classification benchmark, so the OSC\&R results have no external reference point.

\paragraph{Construct validity}
The bootstrap and McNemar tests treat the 451 test segments as independent, but segments from the same source report are correlated, so the reported confidence intervals understate uncertainty at the report level. Flat macro-F1 also counts a BOTH segment predicted as ATTCK as fully wrong, although one of the two target frameworks is correct.

\section{Conclusion and Future Work}
\label{sec:conclusion}
We classified CTI sentences across four classes as ATTCK, OSCR, BOTH, or NON-ATTACK and predicted their technique identifiers, using 2,254 segments from 20 reports with report-grouped splits. The fine-tuned CTI-BERT models (single-head, two-stage, multi-task) reached test macro-F1 between 0.6065 and 0.6392 and did not differ detectably from each other. The QLoRA-adapted Qwen2.5-7B-Instruct model, trained with fixed settings, scored 0.1985, close to the majority-class baseline. For technique identification, the multi-task model reached micro-F1 of about 0.02 and macro-F1 below 0.02. Framework confusion was the largest error type for the CTI-BERT models, led by segments that fit both frameworks. The multi-task CTI-BERT model mislabeled 11 of 270 NON-ATTACK segments but missed 47 of 181 attack segments. Future work will expand the corpus to add training examples for rare techniques and enlarge the test set. Sparse technique labels could be addressed with oversampling~\cite{kim2022comparative}, synthetic sentences~\cite{ruizrodenas2025synthcti}, ontology knowledge~\cite{huang2024mitretrieval}, or retrieval of technique descriptions~\cite{ryan2026evaluating}. A training objective that separates ATT\&CK and OSC\&R evidence could reduce BOTH errors, and a metric that gives partial credit when a BOTH segment is predicted as ATTCK or OSCR would better reflect them. 

\section*{Declaration of Generative AI Technologies}
During manuscript preparation, the authors used ChatGPT only to improve flow, grammar, and clarity. The tool was not used to generate technical content, synthesize citations, or verify experimental facts. The authors reviewed and verified all outputs and take full responsibility for the final manuscript.

\bibliographystyle{ACM-Reference-Format}
\bibliography{references}

\end{document}